\documentclass{article}

 \usepackage[preprint]{neurips_2026}

\usepackage[utf8]{inputenc} 
\usepackage[T1]{fontenc}    
\usepackage{hyperref}       
\usepackage{url}            
\usepackage{booktabs}       
\usepackage{amsfonts}       
\usepackage{nicefrac}       
\usepackage{microtype}      
\usepackage{xcolor}         
\usepackage{amsmath}
\usepackage{graphicx}
\usepackage{multirow}
\title{LH-AVLN: A Benchmark for Long-Horizon Audio-Visual-Language Navigation}

\author{
\begin{tabular}{c}
Rufeng Chen$^{1}$, Yue Chang$^{1}$, Zili Shao$^{1}$, Zhaofan Zhang$^{1}$, \\
Li Chen$^{1}$, Hechang Chen$^{2}$, Hui Xiong$^{1}$, Sihong Xie$^{1}$ \\[0.5em]
$^{1}$\,The Hong Kong University of Science and Technology (Guangzhou) \\
$^{2}$\,Jilin University \\[0.25em]
\texttt{rchen514@connect.hkust-gz.edu.cn}
\end{tabular}
}

\begin{document}

\maketitle

\begin{abstract}
Embodied navigation is moving toward long-horizon missions, yet existing long-horizon benchmarks are largely acoustically silent, and audio-visual navigation tasks typically focus on a single goal. 
We introduce LH-AVLN, a benchmark for Long-Horizon Audio-Visual-Language Navigation that combines multi-goal mission execution, heterogeneous goal specifications, and persistent spatialized acoustic cues. 
In LH-AVLN, an agent receives a global mission of two to four goals specified by category, language description, or reference image, and navigates with RGB-D observations, pose, and binaural audio in indoor 3D environments. 
The benchmark supports both ordered and unordered missions, where alternating goal-associated sounds can guide non-line-of-sight search but may also become distractors as mission progress changes. 
We further develop PAG-Nav, a training-free reference agent that maintains a temporal uniform semantic map and performs progressive goal-state planning, using sound for search while reserving completion for visual-semantic verification. 
Experiments show that existing vision-language, memory-based, and audio-visual agents struggle to complete full LH-AVLN missions, and that PAG-Nav provides a stronger diagnostic baseline while leaving substantial room for future progress.

\end{abstract}

\begin{figure}
    \centering
    \includegraphics[width=1.0\linewidth]{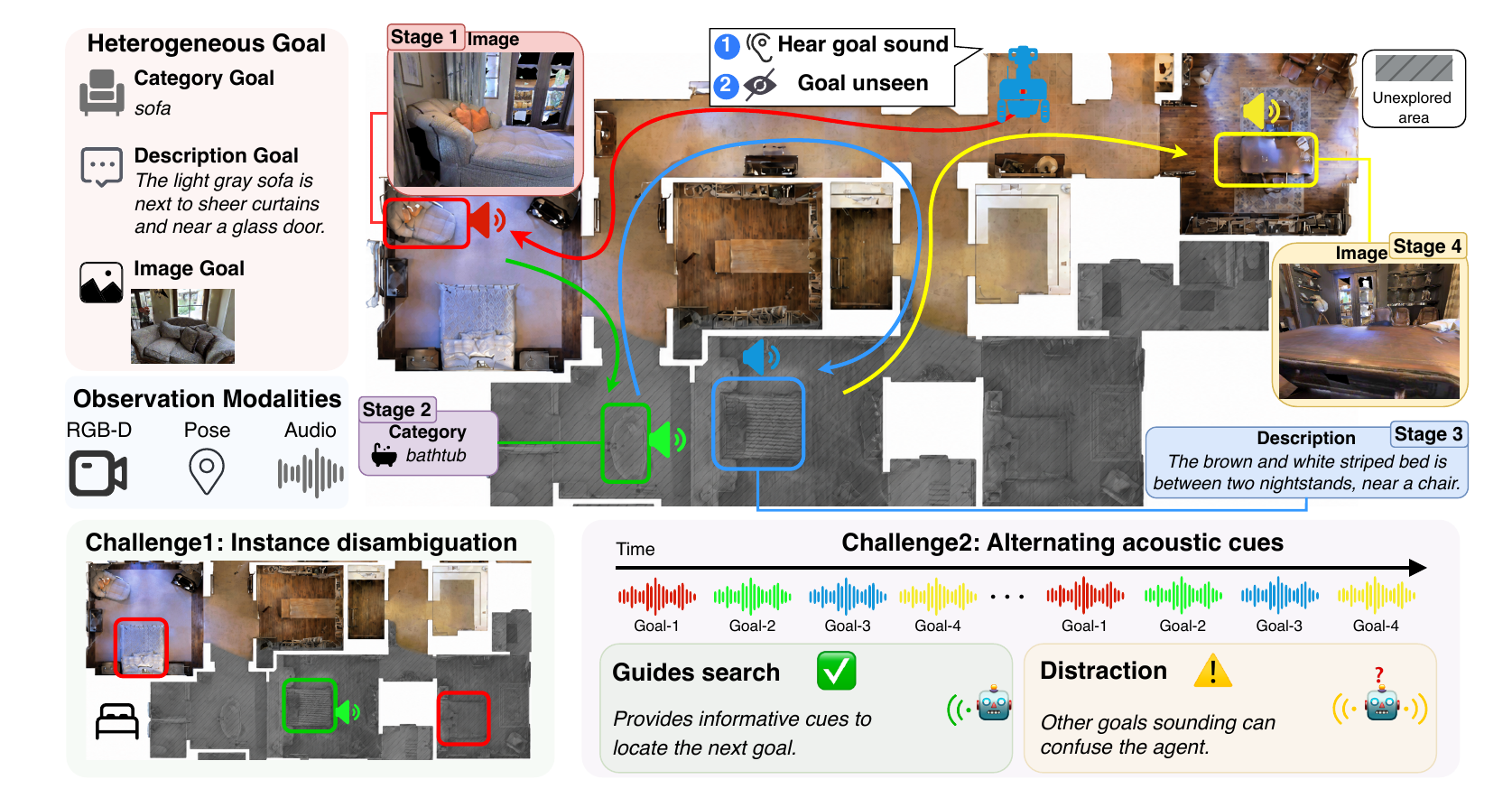}
    \caption{
    {Overview of LH-AVLN.}
    An agent receives a global task with heterogeneous goals specified by category, language description, or reference image, and navigates with RGB-D observations, pose, and binaural audio. 
    Goal-associated sounds can guide non-line-of-sight search, but alternating cues from unfinished goals may become distractors as the mission state changes. 
    LH-AVLN requires target-instance disambiguation, persistent cross-modal memory, and visual-semantic verification under both ordered and unordered multi-goal missions.
    }
    \label{fig:benchmark_overall}
\end{figure}

\section{Introduction}
Embodied navigation is moving beyond short, single-objective episodes. 
Early vision-and-language navigation benchmarks typically evaluate route following or remote object grounding as a single-objective task~\cite{R2R_2018, qi2020reverie, chen2026psg}. 
Recent multi-goal and lifelong benchmarks extend this setting to longer tasks over different modalities (e.g., object, description, image) targets, making scene memory and task-progress tracking central challenges~\cite{wani2020multion, khanna2024goat, song2024towards}. 
Yet these benchmarks largely remain acoustically silent. 
As a result, agents can only collect goal-related information from the current visual field or from previously built visual-semantic memory.

Sound changes this information structure. 
Unlike vision, spatial audio can provide directional cues about objects or events beyond the agent's current field of view, making it useful for searching occluded or unexplored regions~\cite{chen20soundspaces, chen2022soundspaces}. 
This property has motivated audio-visual navigation tasks in which agents localize sound sources, navigate to semantically specified sounding objects, or operate under moving and noisy acoustic conditions~\cite{chen2021semantic, paul2022avlen, zeng26semantic}. 
However, these tasks typically keep the acoustic cue tied to a single currently relevant objective. 
The agent hears where to go, but it does not need to decide which unfinished goal the sound should support, whether the sound has become irrelevant, or whether it conflicts with the current mission state.

In long-horizon audio-visual-language navigation, acoustic cues are not fixed targets to follow. 
They are progress-dependent information. 
A sound may provide a useful spatial prior for an unfinished goal, but it does not identify the corresponding goal, category, or instance. 
The same cue may also become irrelevant or misleading after the agent changes subtask or completes a goal \cite{fan2026navla, shi2025towards, younes2023catch}. 
This makes acoustic grounding fundamentally different from single-objective audio navigation: the agent must reason about both the physical source of a sound and its task-level relevance.
The central challenge is therefore task-conditioned cue grounding rather than generic multimodal fusion. 
An agent must associate acoustic cues with candidate visual objects, suppress cues that are irrelevant to the current goal state, carry useful audio-visual associations across goal transitions, and visually verify the target instance before submission.
Figure~\ref{fig:benchmark_overall} illustrates these two coupled difficulties: acoustic cues can guide non-line-of-sight search, but they must be disambiguated against heterogeneous goal specifications and changing goal-completion states.

To make this problem measurable, we introduce LH-AVLN, a benchmark for Long-Horizon Audio-Visual-Language Navigation (\textbf{LH-AVLN}). 
Rather than adding sound to an existing navigation task, LH-AVLN couples three factors that are usually evaluated separately: heterogeneous goal grounding, long-horizon progress tracking, and spatialized acoustic guidance. 
As summarized in Table~\ref{tab:benchmark_comparison}, existing long-horizon benchmarks largely omit acoustic cues, while audio-visual navigation benchmarks typically focus on a single navigation objective.
Each episode gives the agent a global task with different goals, where each goal is specified by an object category, a language description, or a reference image. 
The agent navigates in indoor 3D environments with RGB-D observations, pose, and binaural audio.
Sound is tied to the mission rather than to a single static target. 
Each mission target is associated with a spatialized sound source, and unfinished targets alternately emit acoustic cues until they are completed. 
As a result, audio can guide search beyond the current visual field, but it can also pull the agent toward another unfinished goal when that cue is irrelevant to the current mission state. 
LH-AVLN evaluates this setting under both ordered and unordered tasks.

We establish reference agents and diagnostic evaluations for LH-AVLN. 
Our training-free agent, PAG-Nav, is designed around the core difficulty of goal-state-aware cue grounding: it maintains persistent geometric, semantic, and acoustic memory, uses audio to guide exploration, and reserves final goal completion for visual-semantic instance verification. 
We further compare against existing vision-language, memory-based, and audio-visual navigation agents to test whether capabilities developed in prior settings transfer to LH-AVLN. 
Experiments show that these agents struggle to complete full missions, even when they occasionally reach individual goal regions. 
This reveals several open challenges: rejecting task-irrelevant sounds, verifying the correct instance, tracking goal progress, and managing long-horizon decisions.

\begin{table*}[t]
\centering
\caption{
Long-horizon and multi-goal benchmarks usually omit acoustic cues, while audio-visual navigation benchmarks typically evaluate a single navigation objective. 
LH-AVLN combines multi-stage task execution, heterogeneous goal specifications, and alternating goal-associated acoustic cues. 
Cat., Txt., and Img. denote category, text, and image goals.
}
\label{tab:benchmark_comparison}
\resizebox{\textwidth}{!}{
\begin{tabular}{lccccc}
\toprule
\textbf{Benchmark}
& \textbf{Task}
& \textbf{Goal Modality}
& \textbf{Acoustic } 
& \textbf{Avg. Task Step}  
& \textbf{Task Num}
\\
\midrule

R2R \cite{R2R_2018}
& Step-by-step
& Txt.
& None 
& <8
&  21567
\\

REVERIE \cite{REVERIE_2020}
& Obj Nav
& Txt.
& None 
& <8
&  21702
\\

VLN-CE \cite{vln_ce_2020}
& Stey-by-step
& Txt.
& None 
& 55.88
&  4475
\\

MultiON \cite{wani2020multion}
& Iterative
& Cat.
& None
& -
& -\\

GOAT-Bench \cite{khanna2024goat}
&  Iterative 
& Cat./Txt./Img.
& None 
& -
&  725360
\\

LHPR-VLN \cite{song2024towards}
& Multi-stage 
& Txt.
& None 
& 216
&  3260
\\

CoNavBench \cite{wang2026conavbench}
&  Multi-agent
& Txt.
& None 
& -
& 4048 
\\

BeDAViN \cite{shi2025towards}
& Obj Nav
& Txt.
&  Multi-source 
& -
& -
\\

MINav \cite{fan2026navla}
& Obj Nav
& Cat./Txt./Img.
& Single-source 
& -
& -
\\

\midrule

\textbf{LH-AVLN (Ours)}
& \textbf{Multi-stage}
& \textbf{Cat./Txt./Img.}
& \textbf{Multi-source} 
& 172.17
& 156550 
 \\

\bottomrule
\end{tabular}
}
\end{table*}

\section{Related Works}

\paragraph{Long-Horizon and Multi-Goal Navigation.}
Embodied navigation benchmarks have progressively increased the temporal horizon and semantic complexity of navigation tasks. 
Early vision-and-language navigation benchmarks such as R2R \cite{R2R_2018} and VLN-CE \cite{vln_ce_2020} evaluate instruction following along a single route, while REVERIE \cite{qi2020reverie} additionally requires grounding a referred object at the destination. 
MultiON \cite{wani2020multion} moves from single-goal navigation to sequential object search, making semantic map memory and observation reuse central to task completion. 
GOAT-Bench \cite{khanna2024goat} further studies lifelong navigation with goals specified by category, language, or image, and evaluates both modular and learned agents with explicit or implicit memory. 
Recent benchmarks extend long-horizon navigation through persistent scene interaction, multi-stage instruction following, and collaborative multi-agent execution~\cite{hong2025general, song2024towards, wang2026conavbench}. 
These works motivate LH-AVLN's focus on long-horizon missions, heterogeneous goal specifications, and task-progress tracking. 
However, they primarily evaluate agents in acoustically silent environments. 
As a result, they do not test whether an agent can use non-line-of-sight acoustic cues, associate them with different unfinished goals, or suppress them after the mission state changes.

\paragraph{Audio-Visual Navigation.}
Audio-visual navigation studies how spatial sound can guide embodied agents beyond the current visual field. 
SoundSpaces introduced realistic audio simulation for indoor 3D environments and enabled tasks such as AudioGoal navigation, where an agent localizes a sounding target from binaural audio~\cite{chen20soundspaces, chen2022soundspaces}. 
Subsequent work extends this setting from source localization to semantic and language-conditioned navigation, where agents use sound together with visual observations and goal specifications to find target objects or events~\cite{chen2021semantic, paul2022avlen, fan2026navla, zeng26semantic}. 
Other benchmarks study more challenging acoustic conditions, including moving sources, unheard sounds, background noise, and multiple sound sources~\cite{younes2023catch, shi2025towards}. 
These works show that audio can provide useful spatial cues when visual evidence is missing or delayed. 
Yet most formulations still center on one currently relevant navigation objective. 
The acoustic cue either directly defines the goal or remains tied to the same target throughout the episode. 
LH-AVLN instead studies persistent goal-associated sounds in a multi-goal mission, where the relevance of a cue changes with task progress and successful completion requires cross-goal memory, distractor rejection, and visual-semantic instance verification.

\section{The LH-AVLN Benchmark}

\subsection{Task Definition}

LH-AVLN evaluates embodied agents in partially observable indoor 3D environments with long-horizon tasks, heterogeneous goal specifications, and target-associated spatialized audio. 
Each episode defines a global mission $\mathcal{G}=\{g_i\}_{i=1}^{N}$, where $N\in\{2,3,4\}$. 
The mission is given to the agent at the beginning of the episode. 
Each goal is represented as $g_i=(m_i, q_i)$, where $m_i\in\{\texttt{Category}, \texttt{Description}, \texttt{Image}\}$ denotes the goal modality and $q_i$ denotes the corresponding goal cue. 
For a category goal, $q_i$ is an object-category label. 
For a description goal, $q_i$ is a natural-language referring expression. 
For an image goal, $q_i$ is a reference image of the target instance. 
Description and image goals are not accompanied by category labels or instance identifiers, so the agent must infer the target semantics from the provided cue.

Each goal $g_i$ is associated with a target instance $y_i$ in the environment. 
The mapping between $g_i$ and $y_i$, together with the identity and location of $y_i$, is used for episode generation and evaluation but is never exposed to the agent. 
At each time step $t$, the agent receives an egocentric observation $o_t=(I_t^{rgb}, I_t^{depth}, p_t, a_t)$, where $I_t^{rgb}$ and $I_t^{depth}$ are RGB-D observations, $p_t$ is the agent pose, and $a_t$ is the binaural audio observation. 
The audio is rendered from spatialized sound sources associated with mission targets. 
It provides directional cues toward potential target locations, but does not reveal the source identity, target category, goal modality, or goal index.

LH-AVLN supports ordered and unordered missions. 
In the ordered setting, the agent must complete goals according to a predefined sequence. 
In the unordered setting, the agent may choose the visitation order and must complete all goals efficiently. 
The action space is $\mathcal{A}=\{\texttt{MoveForward}, \texttt{TurnLeft}, \texttt{TurnRight}, \texttt{Submit}, \texttt{Stop}\}$. 
For an episode with $N$ goals, the agent may issue at most $N-1$ \texttt{Submit} actions and one final \texttt{Stop} action. 
A \texttt{Submit} action marks an intermediate goal as completed if the agent is within the success region of a valid target. 
In the ordered setting, the valid target is the next goal in the prescribed sequence. 
In the unordered setting, any unfinished target is valid. 
The final \texttt{Stop} action terminates the episode and succeeds only if the agent is within the success region of the last remaining valid target. 
After a target is completed, its sound source is deactivated. 
The remaining unfinished targets alternately emit spatialized acoustic cues, causing the active sound source to change over time.

\subsection{Benchmark Construction}

\begin{figure}
    \centering
    \includegraphics[width=1.0\linewidth]{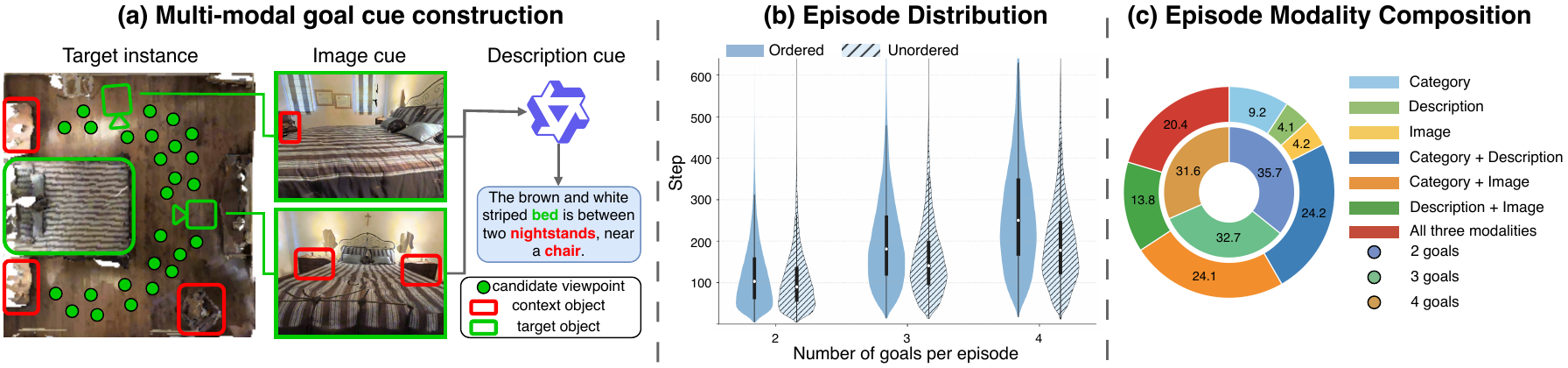}
    \caption{ 
    Benchmark construction and statistics of LH-AVLN.
    (a) Goal cues are constructed from visually validated target instances, including reference images and language descriptions. 
    (b) Reference path lengths cover a broad range of ordered and unordered missions with two to four goals. 
    (c) Episodes contain diverse goal counts and modality compositions.
    }
    \label{fig:Benchmark_construction}
\end{figure}

We construct LH-AVLN to ensure that each episode is navigable, visually verifiable, multimodally grounded, and non-trivial. 
The pipeline first builds a scene-specific pool of valid target instances and goal cues, and then samples long-horizon missions under modality, distance, and reachability constraints. 
Figure~\ref{fig:Benchmark_construction} summarizes the goal-cue construction process and the resulting episode statistics.

\paragraph{Valid target instances.}
For each semantically annotated indoor scene, we enumerate object instances and identify navigable viewpoints from which each instance can be observed. 
We render object-centric views from candidate viewpoints and use an open-vocabulary detector to verify that the target instance is recognizable. 
Instances are removed if they lack a reachable viewpoint or cannot be visually verified from any navigable location. 
The retained instances provide grounded visual evidence for constructing reference images and language descriptions.

\paragraph{Goal cue construction.}
Each valid target instance supports up to three goal cues: its semantic category, a language description, and one or more reference images captured from validated viewpoints. 
When an instance is sampled into an episode, exactly one cue is presented to the agent. 
Description and image goals are not paired with category labels or instance identifiers. 
Each episode contains two to four goals and at least two distinct goal modalities. 
Target categories are unique within an episode, while non-target instances from the same categories may still appear in the scene.

\paragraph{Episode sampling and validation.}
We sample an initial agent pose from the navigation mesh and select a compatible set of target instances from the valid instance pool. 
An episode is retained only if all targets are reachable from the start, each target has at least one valid visual verification viewpoint, and target locations are sufficiently separated in geodesic distance. 
The separation constraint avoids trivial missions and encourages long-horizon exploration. 
We also balance target-instance reuse to improve coverage of the object pool.

\paragraph{Acoustic source assignment.}
Each target instance is assigned a spatialized sound source at its location. 
During an episode, unfinished targets alternately emit acoustic cues. 
Once a target is correctly completed, its sound source is deactivated and removed from the alternation schedule. 
At each time step, binaural audio is rendered according to the relative pose between the agent and the currently active source. 
The acoustic signal provides spatial guidance, but it does not reveal the source category, goal modality, target identity, or goal index.

\paragraph{Reference paths and splits.}
For each accepted episode, we compute geodesic distances between the start position and all selected targets. 
These distances are used to estimate reference path lengths for both ordered and unordered missions. 
We split the benchmark at the scene level, so evaluation measures generalization to unseen environments rather than memorization of training scenes.

\begin{figure}
    \centering
    \includegraphics[width=1.0\linewidth]{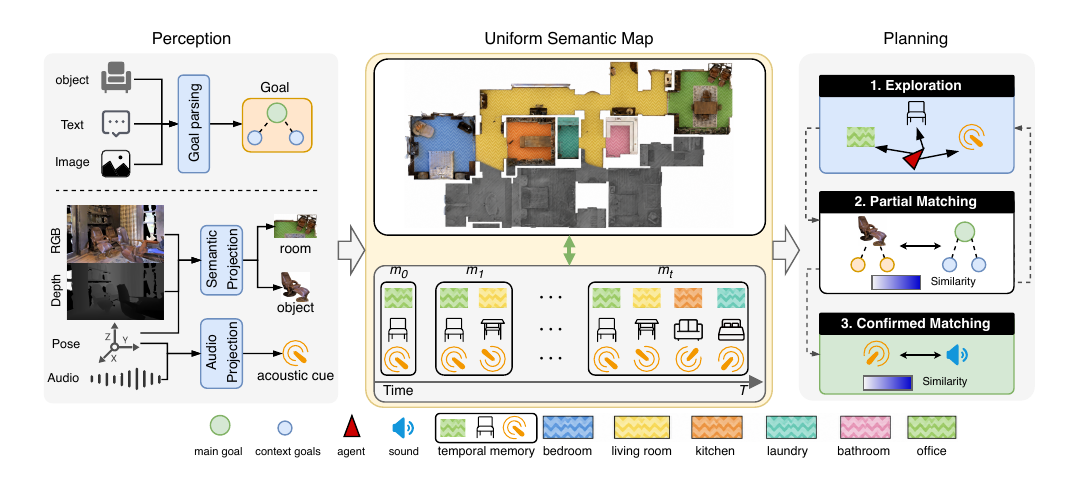}
    \caption{ 
    Overview of PAG-Nav. (Left) The agent receives heterogeneous goal specifications and RGB-D, audio, and pose observations. (Middle) PAG-Nav maintains a uniform semantic map. (Right) A task-conditioned planner switches among exploration, partial matching, and confirmation.
    }
    \label{fig:pag-nav}
\end{figure}

\section{Methods}

We focus on PAG-Nav, a training-free modular agent designed to expose the reasoning structure of LH-AVLN. 
PAG-Nav consists of goal-conditioned perception, a temporal uniform semantic map, and progressive goal-state planning. 
A learning-based policy, MAV-Nav, is described in Appendix~\ref{app:mavnav}.

\subsection{PAG-Nav}

PAG-Nav is a training-free modular agent for LH-AVLN.
As shown in Figure~\ref{fig:pag-nav}, it consists of three components: goal-conditioned perception, a uniform semantic map, and a task-conditioned planner.
The key idea is to use audio for non-line-of-sight search, while reserving visual-semantic matching for target verification.
The semantic map unifies room semantics, object hypotheses, acoustic cues, and temporal memory in a single spatial representation.
Based on this map, the planner switches among exploration, partial matching, and confirmed matching.

\subsubsection{Goal-Conditioned Perception}

At each step $t$, the agent receives an observation $o_t = \left(I_t^{rgb}, I_t^{depth}, p_t, a_t\right),$ together with a mission $\mathcal{G} = \{g_i\}_{i=1}^{N}.$
Each goal $g_i=(m_i,q_i)$ is first encoded into a query embedding $ z_i = \phi_{m_i}(q_i),$ where $\phi_{m_i}$ denotes the modality-specific encoder for category, text, or image goals.

The perception module projects the current observation into three types of map-level cues:
room cues, object cues, and acoustic cues. 
Formally, we write
\[
\Psi_t = \mathcal{P}(o_t, \mathcal{G})
      = \left(\Psi_t^{room}, \Psi_t^{obj}, \Psi_t^{aud}\right),
\]
where $\Psi_t^{room}$ denotes room-semantic predictions, $\Psi_t^{obj}$ denotes goal-relevant object hypotheses, and $\Psi_t^{aud}$ denotes acoustic localization cues estimated from binaural audio.

\subsubsection{Temporal Uniform Semantic Map}

The uniform semantic map is the temporal memory in the middle of Figure~\ref{fig:pag-nav}. 
It accumulates task-relevant room, object, and audio information over the whole episode, rather than representing only the current frame.

At time step $t$, PAG-Nav updates the map as
$
\mathcal{M}_t =
\mathcal{U}\left(
\mathcal{M}_{t-1},
\Psi_t^{room},
\Psi_t^{obj},
\Psi_t^{aud},
p_t,
\mathcal{S}_t
\right),
$
where $\Psi_t^{room}$, $\Psi_t^{obj}$, and $\Psi_t^{aud}$ are the projected room, object, and acoustic cues, $p_t$ is the agent pose, and $\mathcal{S}_t$ is the mission state.

We write the memory as
$
\mathcal{M}_t = (R_t, O_t, A_t).
$
Here, $R_t$ stores room-level semantic context, $O_t$ stores persistent object hypotheses and their goal associations, and $A_t$ stores acoustic memory. 
The key point is that all three are temporal memories in a shared spatial frame.

Room memory $R_t$ provides semantic priors for search. 
Object memory $O_t$ maintains candidate-goal associations over repeated observations. 
Audio memory $A_t$ has two complementary forms:
$
A_t = (A_t^{int}, A_t^{bear}).
$
The intensity memory accumulates local acoustic strength,
$
A_t^{int}(\ell,x_t)
=
\max\left(A_{t-1}^{int}(\ell,x_t), \rho_t\right),
$
while the bearing memory stores directional sound observations,
$
A_t^{bear}(g_i)
=
A_{t-1}^{bear}(g_i)
\cup
\{(p_t,\beta_t,\rho_t)\}.
$

These two audio memories play different roles. 
Bearing memory guides exploration before a reliable object is found. 
Intensity memory supports candidate confirmation after an object hypothesis is available. 
Thus, the uniform semantic map converts partial observations into persistent room-object-audio associations for the planner.

\subsubsection{Progressive Goal-State Planning}
Given the temporal uniform semantic map, PAG-Nav selects a planning stage according to the grounding status of the active goal. 
If no reliable object candidate has been found, the agent performs cue-guided search. 
If a plausible candidate exists but has not been verified, the agent enters candidate grounding. 
If the candidate has been confirmed, the agent performs instance verification before submission:
\[
\pi_t =
\begin{cases}
\texttt{Search}, & \text{no reliable candidate},\\
\texttt{Ground}, & \text{plausible candidate},\\
\texttt{Verify}, & \text{confirmed candidate}.
\end{cases}
\]

\paragraph{Cue-Guided Search.}
When no reliable object candidate is available, PAG-Nav searches over representative points sampled from the current frontier set. 
Let $\mathcal{F}_t=\{x_k\}$ denote the reachable frontier representatives. 
Each point is scored using goal-conditioned semantic priors and acoustic direction:
\[
S_t^{search}(x_k,g)
=
\lambda_r P_{room}\!\left(g \mid r(x_k)\right)
+
\lambda_o P_{obj}\!\left(g \mid \mathcal{N}_{obj}(x_k)\right)
+
\lambda_a s_t^{aud}(x_k,g),
\quad x_k\in\mathcal{F}_t .
\]
Here, $r(x_k)$ is the room type estimated around frontier $x_k$, and $\mathcal{N}_{obj}(x_k)$ denotes the observed object context near that frontier. 
The terms $P_{room}$ and $P_{obj}$ are semantic relevance priors, similar in spirit to the commonsense room-object and object-object associations used in semantic-graph navigation methods such as SG-Nav. 
The audio term $s_t^{aud}$ measures the alignment between the frontier direction and the current acoustic bearing memory. 
PAG-Nav selects
\[
x_t^\star
=
\arg\max_{x_k\in\mathcal{F}_t}
S_t^{search}(x_k,g).
\]
This stage uses semantic priors and sound direction only to decide where to search next. 
It does not create a target match or trigger completion.

\paragraph{Candidate Grounding.}
When the search stage reveals plausible object candidates, PAG-Nav switches from frontier-level search to instance-level grounding. 
Let $\mathcal{O}_t(g)\subseteq O_t$ denote the candidate objects associated with goal $g$ in the temporal object memory. 
For each candidate $o_j\in\mathcal{O}_t(g)$, PAG-Nav computes a grounding score:
\[
S_t^{ground}(o_j,g)
=
\lambda_v s_t^{vis}(o_j,g)
+
\lambda_c P_{ctx}\!\left(g \mid o_j, R_t, O_t\right)
+
\lambda_a s_t^{int}(o_j,g).
\]
Here, $s_t^{vis}$ measures visual-semantic compatibility between the goal cue and the object candidate, $P_{ctx}$ is a commonsense contextual prior from nearby room and object semantics, and $s_t^{int}$ measures consistency with the audio-intensity memory. 
The selected candidate is
\[
o_t^\star
=
\arg\max_{o_j\in\mathcal{O}_t(g)}
S_t^{ground}(o_j,g).
\]
PAG-Nav then navigates toward a viewpoint that can improve the candidate grounding. 
This stage turns a map-level hypothesis into an instance-level candidate, but the object is not yet submittable.

\paragraph{Instance Verification.}
After a candidate has been sufficiently grounded, PAG-Nav enters instance verification. 
The agent navigates to a valid observation pose around the selected candidate $o_t^\star$ and checks whether it can be safely submitted. 
Submission is allowed only when the candidate passes the final verification gate:
\[
\operatorname{Submit}_t
=
\mathbb{I}
\left[
\mathcal{V}_t(o_t^\star,g)
\land
\|p_t-p(o_t^\star)\|\le \delta_{sub}
\land
\angle(h_t,o_t^\star)\le \theta_{sub}
\right],
\]
where $\mathcal{V}_t(o_t^\star,g)$ denotes the visual-semantic verification result, $p_t$ and $h_t$ are the agent position and heading, and $\delta_{sub}$ and $\theta_{sub}$ are the submit distance and heading thresholds. 
This stage separates grounding from commitment. 
Acoustic cues can help the agent reach a plausible candidate, but the final \texttt{Submit} action is triggered only after visual-semantic verification at a proper pose.

\section{Experiments}
We evaluate LH-AVLN to answer three questions: 
(1) how challenging long-horizon audio-visual-language navigation is for existing methods; 
(2) whether spatial audio improves long-horizon goal search beyond vision-language observations alone; and 
(3) how goal ordering, multimodal goal specifications, and persistent memory affect task completion. 
We report results under both ordered and unordered mission settings, where agents must complete two to four heterogeneous goals using RGB-D observations, pose, binaural audio, and the global mission specification.

\textbf{Experimental Setup}
We build LH-AVLN on SoundSpaces 2.0 \cite{chen2022soundspaces} using Matterport3D \cite{chang2017matterport3d} scenes with spatialized binaural audio. 
The RGB-D and audio sensors are mounted at 1.0 m height. 
A target is counted as found if the agent executes Submit or Stop within $2$ m of the target. 
For an episode with ($N$) goals, the agent may issue at most ($N-1$) Submit actions and one final Stop, which completes the last goal and terminates the episode.

\begin{table*}[t]
\centering
\caption{Performance comparison under ordered and unordered goal settings.}
\label{tab:main_result}
\resizebox{\textwidth}{!}{
\begin{tabular}{llc*{10}{c}}
\toprule
\multirow{2}{*}{\textbf{Method}}
& \multirow{2}{*}{\textbf{Type}}
& \multirow{2}{*}{\textbf{Audio}}
& \multicolumn{5}{c}{\textbf{Ordered}}
& \multicolumn{5}{c}{\textbf{Unordered}} \\
\cmidrule(lr){4-8}
\cmidrule(lr){9-13}
&
&
& \textbf{SR$\uparrow$}
& \textbf{ISR$\uparrow$}
& \textbf{OSR$\uparrow$}
& \textbf{SPL$\uparrow$}
& \textbf{SoftSPL$\uparrow$}
& \textbf{SR$\uparrow$}
& \textbf{ISR$\uparrow$}
& \textbf{OSR$\uparrow$}
& \textbf{SPL$\uparrow$}
& \textbf{SoftSPL$\uparrow$} \\
\midrule

MTU3D
& Training-free
& --
& 1.7 & 15.5 & 47.6 & 1.5 & 8.8
& 2.1 & 17.3 & 41.3 & 1.1 & 9.8 \\

3D-Mem
& Training-free
& --
& 0.0 & 11.6 & 23.7 & 0.0 & 8.6
& 1.0 & 14.8 & 28.0 & 0.7 & 9.2 \\

SAVI
& Pretrain
& \checkmark
& 0.0 & 0.1 & 34.4 & 0.0 & 0.1
& 0.0 & 0.7 & 35.1 & 0.0 & 0.5 \\

Goat-bench
& Training-free
& --
& 0.0  & 7.9 & 9.1 & 0.2 & 7.9
& 0.2 & 8.9 &9.2 & 0.2 & 8.9 \\

SCOPE
& Training-free
& --
& - & - & - & - & -
& - & - & - & - & -\\

\textbf{PAG-Nav}
& Training-free
& \checkmark
& \textbf{2.3} & \textbf{17.5} & \textbf{66.5} & \textbf{1.8} & \textbf{12.9}
& \textbf{3.1} & \textbf{31.5} & \textbf{55.6} & \textbf{1.8} & \textbf{14.8} \\

\bottomrule
\end{tabular}
}
\end{table*}

\textbf{Baselines}
We compare PAG-Nav with six representative baselines under both ordered and unordered settings. 
MTU3D \cite{zhu2025move}, 3D-Mem \cite{yang20253d}, Goat-Bench \cite{khanna2024goat}, and SCOPE \cite{wang2026expand} are training-free RGB-D navigation agents that support category, description, and image goals, serving as vision-language-only baselines without acoustic input. 
SAVI \cite{chen2021semantic} and AVLen \cite{paul2022avlen} are pretrained audio-visual navigation agents that use RGB-D, pose, and audio observations, but do not natively support heterogeneous goal specifications. 
These baselines allow us to evaluate whether existing memory-based, vision-language, and audio-visual navigation methods can transfer to long-horizon multi-goal navigation. 
PAG-Nav is our training-free omni-modal baseline, which uses RGB-D, pose, binaural audio, and all three goal modalities.

\subsection{Main Result}
We first evaluate whether existing embodied navigation agents can directly handle LH-AVLN.
Table \ref{tab:main_result} shows that all baselines obtain very low full-task success, indicating that long-horizon audio-visual-language navigation is substantially harder than single-goal or acoustically silent navigation. 
Vision-language agents can occasionally reach or submit partial goals, but their lack of acoustic input limits efficient non-line-of-sight search.
Pretrained audio-visual agents achieve non-trivial oracle success, suggesting that audio helps agents approach sounding regions, but their lack of heterogeneous goal grounding prevents reliable task completion.
PAG-Nav performs best in both ordered and unordered settings, showing that persistent geometric, semantic, acoustic, and goal-progress memory is important for long-horizon multi-goal navigation.

\section{Conclusion}

We introduced LH-AVLN, a benchmark for Long-Horizon Audio-Visual-Language Navigation that combines multi-goal mission execution, heterogeneous goal specifications, and persistent spatialized acoustic cues in indoor 3D environments. 
Unlike prior audio-visual navigation settings that typically bind sound to a single objective, LH-AVLN requires agents to reason about how acoustic cues relate to unfinished goals, changing mission states, and visually grounded target instances. 
We also developed PAG-Nav, a training-free reference agent that uses goal-conditioned perception, a temporal uniform semantic map, and progressive goal-state planning to treat sound as a search cue while reserving completion for visual-semantic verification. 

\bibliographystyle{unsrt}
\bibliography{reference}

\newpage
\appendix
\section{Appendix}
\label{app:mavnav}
\subsection{Method}

Our method consists of two components: \textbf{MAV-Nav} (Multi-goal
Audio-Visual Navigation), an audio-visual actor-critic policy for lifelong
embodied navigation, and a \textbf{Decoupled Target-Routing Curriculum
(DTR curriculum)} for training.

\subsubsection{MAV-Nav Policy}

SAVN-CE provides a memory-based backbone built around a scene memory
Transformer-style state encoder. This encoder stores external memory over
historical state features and attends to this memory when computing the current
policy state. However, SAVN-CE was originally designed for single-target
audio-visual navigation with simple goal specifications, and cannot directly handle our
lifelong navigation setting, where the agent must reason over multiple
multimodal goals, track their completion status, and make task-level decisions
such as when to submit or stop.

To address these limitations, we propose \textbf{MAV-Nav}, an audio-visual
actor-critic policy for lifelong navigation. MAV-Nav keeps the SAVN-CE
scene-memory backbone, but extends the policy
interface from a single target to a lifelong set of multimodal goals. The main
additions are multimodal goal conditioning, goal-status modeling, a lightweight
selector for pending-goal routing, and task-level decisions for submitting a
found goal or ending an episode.

\paragraph{Multimodal lifelong policy.}
Target images and target text are encoded by a frozen CLIP ViT-B/32 into a
shared semantic embedding space, while binaural audio is converted into a
spatial audio feature map and encoded by a CNN. The resulting multimodal state
feature is passed to the scene memory Transformer together with historical
memory tokens, yielding the policy representation used for PPO training and
inference. To condition this representation on the appropriate goal at each
step, MAV-Nav uses an explicit target-routing module.

\paragraph{Target routing.}
When multiple goals remain unfinished, the policy must decide which goal should
condition the current behavior. We therefore add a lightweight goal selector on
top of the shared policy state. Given the memory-aware policy representation,
the encoded multimodal goal embeddings, and the pending/completed goal mask, the
selector scores unfinished goals and produces a selected goal distribution.
The selected goal is used to route goal-conditioned features and suppress
completed or inactive goals before action prediction.

\subsubsection{Decoupled Target-Routing Curriculum}

Training MAV-Nav directly with PPO in the full lifelong setting is
difficult. Memory-based audio-visual policies such as SAVN-CE already require
careful optimization in the single-target setting. Our lifelong task increases
the training difficulty because the policy must solve several coupled problems
at once: it has to explore over a much longer horizon, decide which goal is
currently relevant, remember which goals have already been completed, and learn
when a task-level submit action is valid. Both target selection and submit
timing are hard to learn from sparse rewards: the policy rarely receives a
clear signal about which pending goal should be pursued, and most submit
attempts are uninformative or penalized unless the agent has first reached a
valid goal region. As a result, direct PPO can easily fall into degenerate
behaviors, such as submitting after only a few steps or navigating for the
entire episode without submitting.

To avoid asking PPO to discover exploration, target routing, goal reaching, and
submit timing simultaneously, the DTR curriculum separates target selection
from navigation learning in the early stages.
First, during the DAgger warm start, we use a teacher selector to provide the
current target and train the downstream policy for path following, goal
reaching, and correct submit behavior. This makes the rare submit action
observable during early training without requiring the policy to learn target
routing at the same time. Next, we train the policy with PPO under a simplified
two-goal lifelong setting, where the number of goals, goal distance, and submit
budget are controlled to reduce exploration difficulty and credit assignment
complexity. Once stable reach-and-submit behavior has emerged, we extend PPO
training to the full lifelong setting, using the complete goal set and a longer
episode horizon to expand the exploration range toward the target task. In
parallel with this staged policy training, the selector is trained with
goal-status masks and teacher-selected targets before being coupled with the
navigation policy, and later learns to score pending goals and route the
selected goal representation under the lifelong reward.

\subsection{Experimental Settings}

\textbf{Environments and Simulator.}
Following the main benchmark setting, we build the MAV-Nav evaluation on
SoundSpaces 2.0 \cite{chen2022soundspaces} using Matterport3D
\cite{chang2017matterport3d} scenes with spatialized binaural audio. The RGB-D
and audio sensors are mounted at 1.0 m height. A target is counted as found if
the agent executes \texttt{Submit} or \texttt{Stop} within $2$ m of the target.
For an episode with $N$ goals, the agent may issue at most $N-1$
\texttt{Submit} actions and one final \texttt{Stop}, which completes the last
goal and terminates the episode.

\textbf{Evaluation Protocol.}
We report results on the validation split with 550 full-goal episodes. Each
episode contains two to four heterogeneous goals and uses the complete lifelong
horizon rather than the simplified two-goal curriculum setting. In the ordered
protocol, the current target is determined by the mission order. In the
unordered protocol, MAV-Nav uses its target-routing mechanism to condition the
policy on pending goals and decide which goal should guide the current behavior.

\textbf{Metrics.}
We report full-task success rate (SR), individual subtask success rate (ISR),
oracle success rate (OSR), SPL, and SoftSPL. Following
Table~\ref{tab:main_result}, all metrics in Table~\ref{tab:mav_nav_appendix_results}
are reported as percentages. SR corresponds to completing the entire lifelong
episode, while ISR measures average goal completion across subtasks. OSR
measures whether the agent reaches goal regions regardless of whether it
correctly executes the corresponding task-level submit or stop action. The gap
between OSR and SR therefore reflects failures in target verification, submit
timing, or mission-level completion rather than pure exploration failure.

\begin{table*}[t]
\centering
\caption{Full-goal validation results of a MAV-Nav checkpoint selected from the
DTR curriculum under ordered and unordered goal protocols. All results are
evaluated on 550 validation episodes with OSR enabled. SR, ISR, OSR, SPL, and
SoftSPL are reported as percentages.}
\label{tab:mav_nav_appendix_results}
\resizebox{\textwidth}{!}{
\begin{tabular}{llc*{10}{c}}
\toprule
\multirow{2}{*}{\textbf{Method}}
& \multirow{2}{*}{\textbf{Type}}
& \multirow{2}{*}{\textbf{Audio}}
& \multicolumn{5}{c}{\textbf{Ordered}}
& \multicolumn{5}{c}{\textbf{Unordered}} \\
\cmidrule(lr){4-8}
\cmidrule(lr){9-13}
&
&
& \textbf{SR$\uparrow$}
& \textbf{ISR$\uparrow$}
& \textbf{OSR$\uparrow$}
& \textbf{SPL$\uparrow$}
& \textbf{SoftSPL$\uparrow$}
& \textbf{SR$\uparrow$}
& \textbf{ISR$\uparrow$}
& \textbf{OSR$\uparrow$}
& \textbf{SPL$\uparrow$}
& \textbf{SoftSPL$\uparrow$} \\
\midrule

\textbf{MAV-Nav}
& DTR
& \checkmark
& \textbf{2.2} & \textbf{18.6} & \textbf{45.9} & 0.5 & 11.9
& \textbf{6.0} & \textbf{35.2} & 42.9 & \textbf{3.6} & \textbf{26.9} \\

\quad w/o Audio
& Ablation
& --
& 0.9 & 8.0 & 33.5 & 0.4 & 4.7
& 1.3 & 17.2 & 35.2 & 0.3 & 9.0 \\

\quad w/o Image/Text
& Ablation
& \checkmark
& 1.1 & 17.8 & 43.3 & \textbf{0.6} & \textbf{12.1}
& 4.5 & 34.9 & \textbf{43.4} & 2.5 & \textbf{27.3} \\

\quad DAgger only
& Imitation
& \checkmark
& -- & -- & -- & -- & --
& -- & -- & -- & -- & -- \\

\quad Direct PPO
& RL
& \checkmark
& -- & -- & -- & -- & --
& -- & -- & -- & -- & -- \\

\bottomrule
\end{tabular}
}
\end{table*}

\paragraph{Discussion.}
Table~\ref{tab:mav_nav_appendix_results} shows that full-goal LH-AVLN remains
difficult for the learned policy. The unordered protocol obtains higher SR,
ISR, SPL, and SoftSPL than the ordered protocol. This does
not imply that unordered navigation is intrinsically easier; rather, the
target-routing interface gives the policy more flexibility to exploit whichever
pending goal it can reliably approach and submit. In contrast, the ordered
protocol must follow the prescribed goal sequence, so early failures can block
later progress.

The two protocols have comparable OSR, with ordered evaluation slightly higher.
This indicates that the agent can still reach goal regions under both
protocols, but reaching a region is not sufficient for task completion. The
large OSR--SR gap suggests that the main remaining bottleneck is converting
goal-region visits into correct task-level decisions, especially submit timing
and mission-state management. We therefore report these numbers as full-goal
validation results for MAV-Nav, while keeping earlier two-goal curriculum
evaluations separate because they use a smaller goal set and lower episode
difficulty.


\newpage

\end{document}